\pdfoutput=1

\documentclass[11pt]{article}
\usepackage{authblk}
\usepackage[final]{ACL2023}

\usepackage{times}
\usepackage{latexsym}

\usepackage[T1]{fontenc}

\usepackage[utf8]{inputenc}

\usepackage{microtype}

\usepackage{inconsolata}
\usepackage{graphicx}
\usepackage{float}
\usepackage{colortbl}
\usepackage{xcolor}
\usepackage{multicol}
\usepackage[english]{babel}
\addto\extrasenglish{%
}
\newcommand*{\fullref}[1]{\hyperref[{#1}]{\autoref*{#1} \nameref*{#1}}}
\usepackage{microtype}

\title{Deep Dive into the Language of International Relations: NLP-based Analysis of UNESCO's Summary Records}

\author[1a]{Joanna Wojciechowska}
\author[1a]{Mateusz Sypniewski}
\author[1a]{Maria Śmigielska}
\author[1a]{Igor Kamiński}
\author[2,1a]{\authorcr Emilia Wiśnios}
\author[1b]{Hanna Schreiber}
\author[1b]{Bartosz Pieliński}
\affil[1a]{Faculty of Mathematics, Informatics and Mechanics, University of Warsaw} 
\affil[1b]{Faculty of Political Science and International Studies, University of Warsaw} 
\affil[2]{NASK (National Research Institute)}

\begin{document}
\maketitle
\begin{abstract}
Cultural heritage is an arena of international relations that interests all states worldwide. The inscription process on the UNESCO World Heritage List and the UNESCO Representative List of the Intangible Cultural Heritage of Humanity often leads to tensions and conflicts among states. This research addresses these challenges by developing automatic tools that provide valuable insights into the decision-making processes regarding inscriptions to the two lists mentioned above. We propose innovative topic modelling and tension detection methods based on UNESCO's summary records. Our analysis achieved a commendable accuracy rate of 72\% in identifying tensions. Furthermore, we have developed an application tailored for diplomats, lawyers, political scientists, and international relations researchers that facilitates the efficient search of paragraphs from selected documents and statements from specific speakers about chosen topics. This application is a valuable resource for enhancing the understanding of complex decision-making dynamics within international heritage inscription procedures.
\end{abstract}

\section{Introduction}
The United Nations Educational, Scientific and Cultural Organization (UNESCO) is an international intergovernmental organisation that fosters cooperation in education, science, and culture among its members (currently 194 states). It is the most important universal organisation responsible for promoting and safeguarding cultural heritage, a matter of great concern worldwide. Under the auspices of UNESCO, many international legal agreements were adopted, among them the World Heritage Convention (1972) and the Intangible Cultural Heritage Convention (2003)~\cite{1972_commentary, 2003_commentary}. These conventions established two famous UNESCO heritage lists: the World Heritage List (Convention 1972) and The Representative List of the Intangible Cultural Heritage of Humanity (Convention 2003). The inscriptions of outstanding cultural sites or intangible cultural traditions and practices to these lists (UNESCO heritage lists) shall promote mutual respect and dialogue among states. However, because those inscriptions bring prestige to states having them~\cite{articleSchreiber} and economic boost for communities associated with them~\cite{Bortolotto2020}, there is a lot of competition between states regarding their visibility on the UNESCO heritage  lists~\cite{unesco-tension}. States are prone to inscribe as many of "their" elements on the lists as possible~\cite{Meskell2012}. At the same time, the character of the UNESCO heritage lists, which promotes the common cultural heritage of humanity and the diplomatic character of the decision-making process, creates a situation in which open conflicts are infrequent. Therefore to follow the politics behind the lists, one has to focus on less apparent expressions of disagreements between states -- tensions -- which can be identified in summary records published by UNESCO.  \\ 
Despite accumulating substantial textual data produced from the moment of 
establishment of the UNESCO heritage lists, these documents needed consistent structuring to ensure their analysis using automated and Artificial Intelligence (AI) tools. Using diplomatic language poses unique challenges for machine learning models trained on standard datasets, as it differs significantly from formal texts like Wikipedia or informal such as Twitter. Diplomatic language is known for its diplomatic speech acts, such as hedging, indirectness, rhetorical devices, persuasive techniques, and diplomatic formulas, making it difficult for models to discern the intended meaning. These subtle linguistic nuances and references require a deep understanding of the cultural, political, and historical context in which they are used~\cite{burhanudeen2006diplomatic, topalua2014morphological, pokharel2020diplomatic}. Domain research~\cite{political_language, mining-ethos} has highlighted these challenges and emphasised the urgent need for not yet developed approaches to analysing diplomatic language. 
\paragraph{Tension Operationalisation}\label{sec:tension}
Our research aims to create automatic tools that provide insights into the decision-making processes on the international level by identifying instances of tensions between actors~\cite{unesco-tension}. They are not frequently observed because many state interactions at international level are based on consensus. The diplomatic practice demands that all public discussions be pre-planned and expressed politely. Some things that appear controversial to the untrained eye are sanctioned ways of discussing terminology or procedural issues. There is no actual conflict behind them, although the rhetorical form may suggest it. Tensions are very sporadic moments during discussions when actors express their disagreement with the actions of UNESCO bodies or representatives of other State Parties to UNESCO Conventions. Tension - for the sake of this project - appears when an actor involved in an international decision-making process expresses its opinion on a particular decision or topic that is considered as constituting a threat to their interest or officially promoted set of values. Therefore, to identify tensions on the operational level, one has to reject all controversial issues that are only controversial by their rhetorical form but are focused on purely linguistic, procedural or technical issues. Only then are we left with a specific type of controversial issues -- tensions --  rooted in disagreements related to states' interests and values. 
A sample paragraph from summary records of Intangible Cultural Heritage Committee meeting in 2017 that contains tension is: \\
\emph{The delegation of India congratulated the Evaluation Body for the presentation of its very comprehensive report and for its work, adding that 50 nomination files in one year was no mean feat. However, the delegation noted that there were more cases of referral than it would like to see, and questioned why this was so, especially as Committee Members and States Parties did not have the chance to clarify or to supply additional information that would have improved the process. It referred to the 1972 Convention in which there was a clear window for States Parties to supply additional information that inevitably improved the chance of success and inscription, which was ultimately the objective as this boosted communities back home. The delegation thus recommended that the Convention include a time window during which States Parties could clarify and supply additional information. [...] In this regard, the delegation sought a more in-depth discussion on the issue and stated the case for an open-ended working group of States Parties, also open to Observers, that would bring these recommendations to the next Committee session for adoption, and then on to the General Assembly, which would lead to greater interaction, transparency and dialogue between the Evaluation Body and the States Parties.} \\
A controversial issue is defined in~\cite{controversy-in-news} as one that invokes conflicting sentiments or views, which can be represented by the disparity in volume between two polarities. We decided to base our research and approaches on the results from the previous controversy detection research~\cite{controversy-in-news}, but we narrowed it to tension detection (see above).\\
Studying tensions based on a large corpus of documents stretching from the first World Heritage Committee meeting in 1973 to the present day allows international affairs and political science researchers to analyse what topics for which set of actors have been perceived as threats to their interests and values and how these situations were managed. This data also allows for comparing the political dynamic at UNESCO to discussions at other international organisations and capturing a potential specificity of the organisation's power play focused on preserving humanity's cultural heritage. 

\paragraph{Contributions} The paper's contributions can be summarised as follows:
\begin{enumerate}
    \item Development of a language model that classifies paragraphs by tension using a pre-trained language model. 
    \item Identification and extraction of additional linguistic properties: speaking actors and topics.
    \item Creation of a Graphical User Interface application that enables practitioners and researchers to find paragraphs from the transcripts with desired properties quickly. 
    \item Development of a tool allowing longitudinal studies of tensions in international affairs on the example of one selected international organisation documents (UNESCO).
\end{enumerate}

\section{Dataset preprocessing}\label{sec:dataset}

\paragraph{Fetching documents}\label{par:fetching}
Our dataset was comprised of summary records obtained through web scraping from the World Heritage Convention\footnote{\url{http://whc.unesco.org/en/sessions/}} and Intangible Cultural Heritage Convention\footnote{\url{https://ich.unesco.org/en/decisions/}} websites. Specifically, we collected 98 documents from ordinary and extraordinary World Heritage Committee (WHCOM) sessions and 15 from ordinary and extraordinary Intangible Cultural Heritage Committee (ICHCOM) sessions. They form a complete database of all available summary records from the meetings of these organs of both conventions. Each paragraph in the transcript typically represents an actor's statement, which could be written in direct or reported speech.  \\
The documents were available in both English and French. For our analysis, we focused exclusively on documents written in English. However, it is worth noting that summary records from specific years contained sections written solely in French (see \fullref{par:translate}). In total, our dataset contained roughly 6.3 million words from 113 documents.

\paragraph{Text extraction}\label{par:text_extraction}
The summary record files could be divided into three groups based on how they were created: 
\begin{itemize}
    \vspace{-3mm}
    \item Scans in pdf format.
    \vspace{-3mm}
    \item Scans with a copyable layer of text on top, added with an optical recognition program by the document authors, in pdf format.
    \vspace{-3mm}
    \item Born digital documents in pdf or DOCX/DOC format.
\end{itemize}
\vspace{-3mm}
For the first two, we used the open-source optical character recognition software\footnote{\url{https://github.com/ocrmypdf/OCRmyPDF}} to add our layer of copyable text, as we found our program produced better results than the probably less-modern methods applied by the document authors. The text data from the PDF documents were later extracted using a Python library pdfminer.six\footnote{\url{https://github.com/pdfminer/pdfminer.six}}, and from the DOC/DOCX documents by using python-docx\footnote{\url{https://python-docx.readthedocs.io}}.

\paragraph{Splitting the text into paragraphs} \label{par:split_text}
We used paragraphs as our main unit of text for two reasons. First, paragraphs are natural units of text that often contain cohesive and coherent information. In the case of reported speech in the summary records, one paragraph typically addresses all issues raised by the speaker. In the case of direct speech, the situation is more complex, as one speech may consist of several paragraphs of varying lengths. Joining all paragraphs into a single statement could introduce bias into the model, as very long texts are more likely to be dense. Second, the model input size was limited by the constraints of the RoBERTa model \cite{roberta} that we used as our base architecture. With paragraph splitting, the longest input size was about 800 words.

\begin{figure}[H]
\includegraphics[width=\columnwidth]{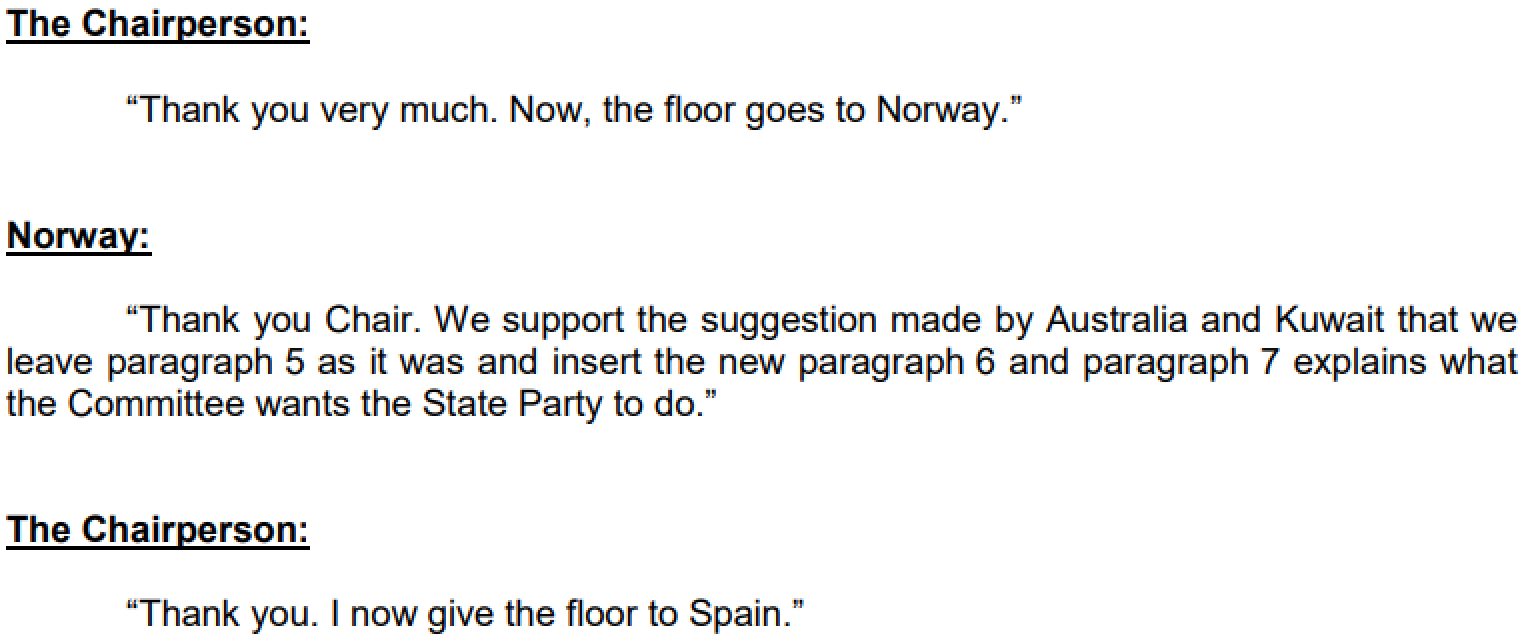}
\centering
\caption{Original paragraphs example}
\label{fig:original_pars}
\end{figure}

The summary record files had no consistent structure, with document elements such as table formatting and ways of using reported and direct speech changing throughout the years, making subdividing the text data into paragraphs more challenging. We solved this problem by using a combination of many different regular expression patterns handcrafted explicitly to detect the various cases of breaks between paragraphs we have analysed. \autoref{fig:original_pars} depicts an example image of the original document structure with \autoref{tab:pars_table} showing the divided paragraphs. Further details on the applied heuristic are presented in \autoref{appendix:splitting_text}.

\paragraph{Spelling correction}\label{par:spelling}
The usage of the optical character recognition program, along with the poor quality of the scanned files, has inevitably introduced some minor errors into the dataset. Moreover, the documents still contained non-text data, such as tables and numbers. We were able to remove some instances of these problems with more regular expressions and tried to fix the spelling errors with many different tools for spelling correction, like Hunspell\footnote{\url{https://github.com/tokestermw/spacy\_hunspell}}, SymSpell\footnote{\url{https://github.com/wolfgarbe/SymSpell}}, or symspellpy\footnote{\url{https://github.com/mammothb/symspellpy}}. However, we found that while these programs did correct some of the errors, they all introduced new errors of their own, and ones that could not be ignored, e.g. changing the word \emph{UNESCO} to \emph{enesco} or \emph{(United Arab) Emirates} to \emph{Pirates}. We ultimately decided not to use a spelling correction tool.

\paragraph{French to English translation}\label{par:translate}
The summary records include transcripts of State Parties using French. 
We decided to translate those parts of our data into English for the controversy detection task. To do so, we used GoogleTranslator~\cite{google_translate} on sentences that were detected to most likely be French by langdetect\footnote{\url{https://github.com/shuyo/language-detection}}. We also removed paragraphs that langdetect did not classify as either English or French, as they were mostly comprised of OCR and poor scan quality artefacts. 

\vspace*{-7mm}
\begin{table}[H]
    \centering\begin{tabular}{|p{70mm}|} 
\hline
 \rowcolor{gray!50}
 \textbf{Extracted paragraphs example.} \\ [0.5ex] 
 \hline
 The Chairperson:

"Thank you very much. Now, the floor goes to Norway." \\ 
 \hline
 Norway:
 "Thank you Chair. We support the suggestion made by Australia and Kuwait that we leave paragraph 5 as it was and insert the new paragraph 6 and paragraph 7 explains what the Committee wants the State Party to do." \\
 \hline
 The Chairperson:
 "Thank you. I now give the floor to Spain." \\
 \hline
\end{tabular}
    \caption{Extracted paragraphs example}
    \label{tab:pars_table}
\end{table}
\vspace*{-5mm}

\begin{figure}[H]
    \centering 
    \vspace*{-5mm}
    \hspace*{-10mm}
    \includegraphics[width=95mm]{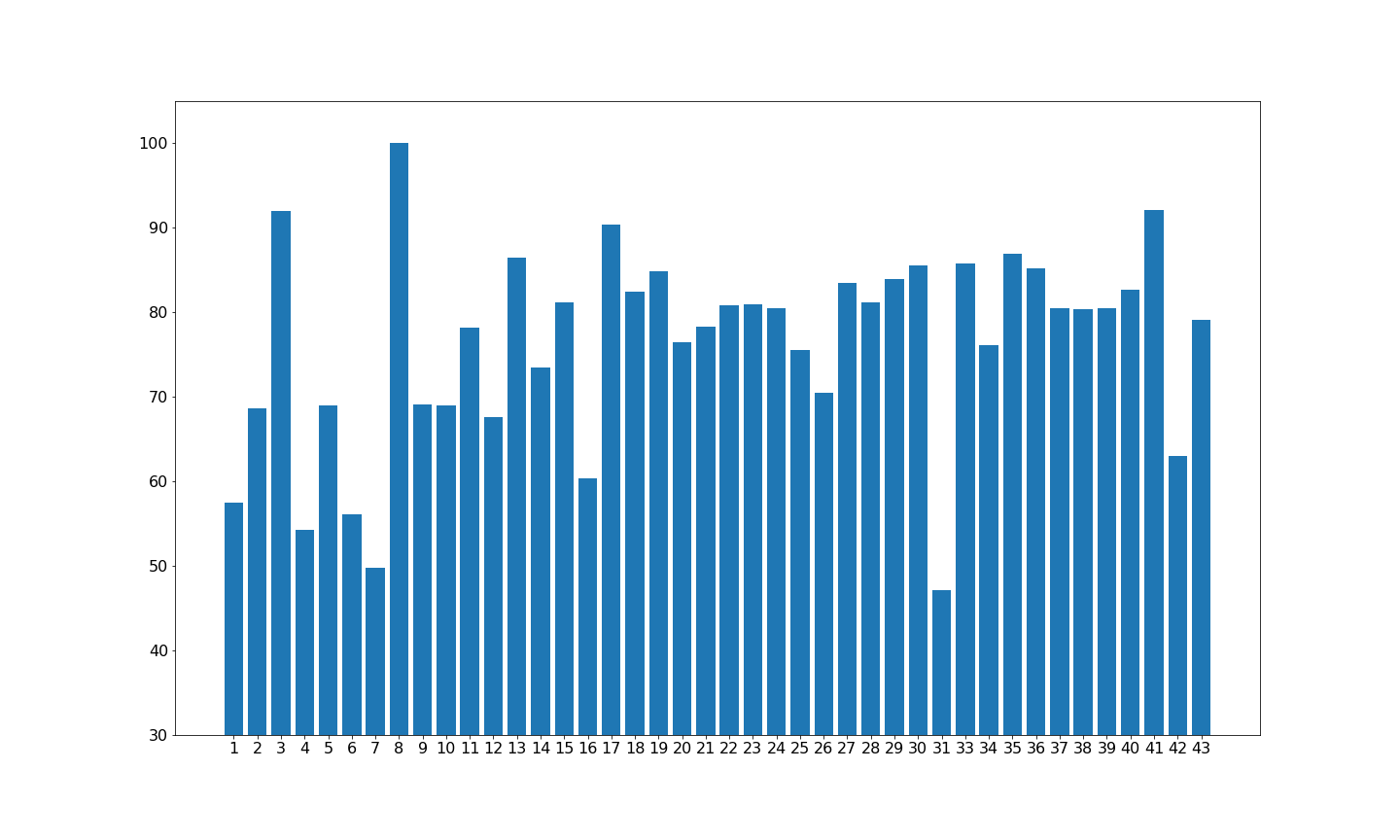} 
    \vspace*{-10mm} 
    \caption{Percentage of paragraphs with identified speakers in each ordinary WHC session.}
    \label{fig:whc_speakers_pars}
\end{figure}

\paragraph{Speaking actor extraction}\label{par:speaking_actor}
A vital feature of the data is that it primarily consists of descriptions of what was said and by whom. Assigning speakers to paragraphs is essential from the perspective of political science research. Speaking actors include individuals with a specific function (\emph{the Chairperson}, \emph{the Rapporteur}), representatives of State Parties to the UNESCO Conventions  (\emph{the Delegation of Turkey}, \emph{the British representative}), representatives of other organisations such as UNESCO advisory bodies or non-governmental organisations. The speaker will only be mentioned by full name in rare cases. The script recognised occurrences of phrases that could be the speaker and assigned the first occurring one as the speaker of the paragraph. The phrases are:
\begin{enumerate}
    \item Specific organisation or role such as \textbf{Chairperson}, \textbf{Rapporteur}, \textbf{ICOMOS} (International Council on Monuments and Sites), \textbf{IUCN} (International Union for Conservation of Nature), \textbf{ICCROM} (International Centre for the Study of the Preservation and Restoration of Cultural Property).
    \item Phrases like `delegation of X', `delegate of X', with `X' replaced with a country name.
\end{enumerate}
The results depicted in \autoref{fig:whc_speakers_pars} show the percentages of paragraphs with detected speakers in each ordinary session of the World Heritage Convention (results for ICH Convention see \autoref{app:ich}) The average results surpass 70\%, particularly in newer documents. It is essential to acknowledge that assigning speakers is not always feasible in every paragraph. Several factors contribute to this limitation. Firstly, certain sections lack explicit speaker attribution as they either provide supplementary information such as lists, introductions or quotes or are part of a larger statement where only the first paragraph contains a speaker phrase. Furthermore, the identification of specific speakers poses challenges when relying solely on regular expressions, especially in cases where individuals are referred to by their full names or when representatives of specific organisations unrelated to UNESCO, such as the Wildlife Conservation Society\footnote{\url{https://www.wcs.org/}}, are mentioned. Moreover, the data quality can impact speaker detection, particularly in older texts where poor data quality becomes prevalent.

\section{Data annotation}\label{par:expert-labelled}
To enable the implementation of the tension detection model, we required labelled text data indicating whether paragraphs contained tension or not. Two domain experts were involved in this process, assigning binary labels to the samples of datasets mentioned in \autoref{sec:dataset}. A label of 1 was assigned if a paragraph contained tension and 0 if it did not. The experts labelled 654 from WHC sessions and 616 from ICHC sessions. After all annotation steps, 321 paragraphs were labelled as ones that contain tension. Details regarding the annotation process are available in the \autoref{appendix:whc_ich_labelled}.

\section{Topic modelling}\label{sec:topic_modelling}
Topic modelling is a text-mining method used to identify and extract hidden topics from large corpora of text data. These topics are usually represented as small sets of keywords or phrases that best capture the topic's semantic meaning~\cite{what-is-topic-modeling}. Historically, the standard approach was to treat the document as a bag of words, disregarding the word order~\cite{blei2003latent}. In recent years, however, there has been a surge in neural network-based topic modelling approaches leveraging pre-trained models, such as BERT~\cite{bert}, following the idea that learned word- and document-level embeddings can provide richer context information than bag-of-words~\cite{topic-modelling-review}. \\ 
To deepen our understanding of the data and as they will be helpful during the building of our application, we used one such powerful topic modelling tool, BERTopic~\cite{bertopic}, to generate a representation of the topics most often brought up in our dataset. BERTopic uses clustering techniques to divide data based on semantic similarity into distinct groups, each constituting a different topic, and then retrieves their keywords and phrases. \\ 
Due to the specific, diplomatic nature of the language used in our dataset, the topics generated by the BERTopic model out-of-the-box could have been better, with most topic key phrases extensive and generic, not describing any meaningful topics. To mitigate this problem, we used the spaCy library~\cite{spacy2} to classify words in our dataset into lexical categories. We removed all but adjectives, nouns, adverbs, and verbs, as we theorised they carried the most semantic meaning. On top of that, we performed stemming~\cite{khyani2021interpretation}. Then, we removed all stopwords and a hand-picked list of overwhelmingly popular words that we did not want to influence the paragraphs' topics, such as \emph{Rapporteur} and \emph{delegate}. We provide the complete list of removed phrases in \autoref{appendix:removed_phrases}. This experiment proved successful; after running BERTopic on the modified paragraphs, we obtained a list of 1024 topics. We performed a human rating of the quality of obtained topics, similar to \cite{NEURIPS2021_0f83556a}. We randomly sampled 100 paragraphs with their topics. Then, we assigned two people to independently rate each paragraph on a scale from 0 to 2, where 0 meant \textit{Not very related}, and 2 meant \textit{Very related}. The average scores for the sampled topics were 1.48 and 1.46. For reference, the scores for topic modeling without text preprocessing were 0.91 and 0.83.


\section{Tension classifier}\label{sec:tension_classifier}
\begin{figure*}[ht]
    \centering
    \vspace*{-10mm} 
    \includegraphics[scale=0.5]{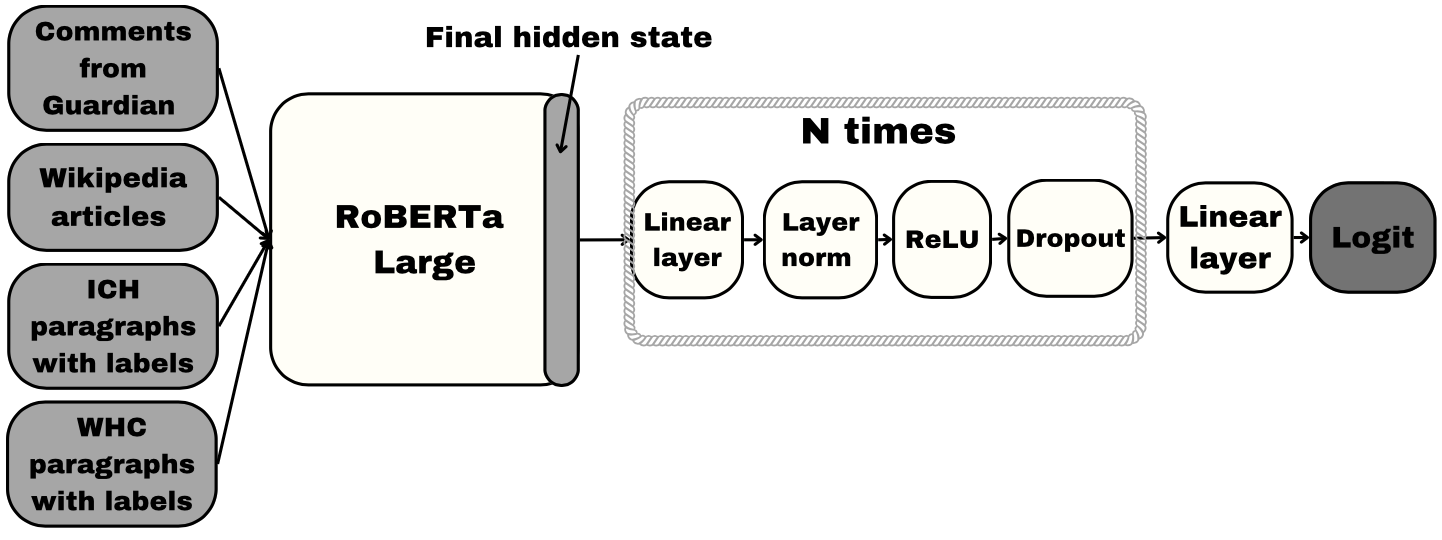}
    \vspace*{-5mm} 
    \caption{Architecture of tension model.}
    \label{fig:model}
\end{figure*}

When developing our initial model, we aimed to perform supervised classification experiments using a pre-trained language model augmented with additional layers. For this purpose, we chose RoBERTa~\cite{roberta}, comparing the results between RoBERTa base and RoBERTa large versions, which differ by the number of parameters and size of a training set. Both of these models are readily available online via Huggingface\footnote{\url{https://huggingface.co/roberta-base}, \url{https://huggingface.co/roberta-large}}. \\ 
Our \emph{tension model} consists of multiple blocks, with each block comprising a Linear Layer, an Activation function - ReLU~\cite{relu}, Layer Norm~\cite{layer_norm}, and a Dropout layer~\cite{dropout}. The final layer of the model consists of a Linear Layer responsible for producing the classification logits. A visual representation of the architecture can be seen in \autoref{fig:model}.

\paragraph{Class imbalance}
Upon intuitively and empirically examining the ICHC and WHC datasets, it becomes evident that a substantial disparity exists between the number of positive and negative samples. This phenomenon, commonly known as class imbalance, is a prevalent challenge in NLP, especially in the context of classification problems~\cite{imbalance-survey}.\\
Two classical approaches are commonly employed to address the class imbalance in datasets. The first approach, known as random undersampling (RUS)~\cite{class_imbalance}, involves randomly removing a selection of majority instances (in our case, negative instances) from the dataset. Although RUS risks discarding potentially valuable data, we empirically decided to drop specific data segments, namely the introductory parts, which we deemed less relevant for our analysis. In our case, we removed 20 paragraphs from the beginning.  
We explored BCEWithLogitsLoss\footnote{\url{https://pytorch.org/docs/stable/generated/torch.nn.BCEWithLogitsLoss.html}}, specifically a parameter \textit{pos\_weight}, which can be used to balance the ratio between classes. During training, we manually selected which values to use.

\paragraph{Active learning}
Active learning, as outlined in~\cite{al-survey}, employs a strategic approach to select the data that should be labelled to maximise the impact on training a supervised model, eg. enhance precision and help with class imbalance.  \\
Rather than randomly selecting a subset of data for manual labelling, we adopted a targeted approach called Uncertainty Sampling~\cite{uncertainty}, where we select samples near the decision threshold, in our case 0.5. We chose 20 samples during each iteration and sought expert annotations for them. Once the paragraphs of interest were labelled, we incorporated these newly labelled samples into the training set.

\subsection{Tension classifier experiments} \label{sec:tension_experiments}
To assess the effectiveness of our model, we utilised three metrics: recall, precision, and accuracy. We conducted a series of experiments to identify the optimal hyperparameters for our model. 
Let us call experiments with additional fine-tuning with datasets based on Guardian and Wikipedia, described in \autoref{sec:related_work}, as experiments with pre-fine-tuning. It is important to note that all our experiments were subjected to fine-tuning with the expert-labelled datasets outlined in \autoref{appendix:whc_ich_labelled}. If pre-fine-tuning was conducted, it was consistently performed before the main fine-tuning process. \\
Throughout the training process, we kept the weights of the RoBERTa model frozen, ensuring that only the weights of the Linear Layers and Layer Norm were subject to gradient updates. Our main goal was to find the best parameters for the number of Linear Layers, Dropout, pos\_weight parameter. Moreover, we wanted to determine which base model we choose (if we need a bigger space of parameters to tune or not) and if we should do pre-fine-tuning.  \\
The splitting ratio of training and test was equal to 8:2 (80\% of samples were used in training, 20\% for testing purposes). The hyperparameters we were looking for were determined based on test set results. The label distributions in the training and evaluation sets are equal to 7:2, with 260 positive and 905 negative in a test set and 65 positive and 227 negative in a training set. \\
Pre-fine-tuning took around 20 epochs, whereas the fine-tuning on datasets described in \autoref{appendix:whc_ich_labelled} took between 6 and 12 epochs. The weight decay parameter was equal to 0.0001, and the learning rate was 0.0005 in all presented experiments.

\paragraph{Pre-fine-tuning} This experiment batch compares results with and without pre-fine-tuning. We suspected that paragraphs using non-diplomatic speech might not be suitable for the inference purpose of our model. On the other hand, we observed that certain expressions, such as \emph{is/go wrong}, \emph{expressed concern}, or \emph{want to discuss/have a debate}, have a universal nature that transcends language boundaries. These expressions often carry implicit meanings and can indicate underlying tensions, regardless of the specific language or cultural context. By incorporating these expressions into the pre-fine-tuning process, our model can benefit from the prior knowledge of their association with tension, thereby enhancing its ability to detect tension in diplomatic discourse. In all experiments, dropout = 0.4, linear blocks = 3, and the base model was RoBERTa large. Results can be seen in \autoref{tab:pre-fine-tuning}.
\begin{table}[th]
\small
\begin{center}
\begin{tabular}{|p{25mm}|p{15mm}|p{10mm}|p{13.5mm}|} 
\hline
 \rowcolor{gray!50}
 \textbf{Description} & \textbf{Precision} & \textbf{Recall} & \textbf{Accuracy} \\ [0.5ex] 
 \hline
 \textit{pos\_weight} = 5 with pre-fine-tuning & 0.82 & 0.52 & 0.54 \\ 
 \hline
 \textit{pos\_weight} = 5 with pre-fine-tuning with removing beginning & 0.81 & 0.44 & 0.45 \\ 
 \hline
 \textit{pos\_weight} = 5 without pre-fine-tuning & 0.79 & 0.5 & 0.52 \\ 
 \hline
 \textit{pos\_weight} = 10 with pre-fine-tuning & 0.79 & 0.72 & 0.72 \\ 
 \hline
 \textit{pos\_weight} = 10 with pre-fine-tuning with removing beginning & 0.75 & 0.60 & 0.64 \\ 
 \hline
 \textit{pos\_weight} = 10 without pre-fine-tuning & 0.82 & 0.52 & 0.57 \\ 
 \hline
\end{tabular}
\caption{Comparision of performance with and without pre-fine-tuning.}\label{tab:pre-fine-tuning}
\vspace{-1.5em}
\end{center}
\end{table}
\paragraph{Dropout} The comparision of different dropouts is in \autoref{tab:dropout}.  All experiments consist of 3 linear blocks and dropout equal 0.4. 

\begin{table}[ht]
\small
\begin{center}
\begin{tabular}{|p{22mm}|p{13.5mm}|p{9mm}|p{13.5mm}|} 
\hline
 \rowcolor{gray!50}
 \textbf{Description} & \textbf{Precision} & \textbf{Recall} &  \textbf{Accuracy} \\ [0.5ex] 
 \hline
 Dropout = 0 & 0.75  &    0.64   &   0.68  \\ 
 \hline
 Dropout = 0.2 & 0.82   &   0.52   &   0.55  \\ 
 \hline
 Dropout = 0.4 & 0.82   &    0.58  &    0.61 \\ 
 \hline
 Dropout = 0.6 & 0.82   &   0.53   &   0.55  \\ 
 \hline
\end{tabular}
\caption{Comparision of performance with different dropouts of the tension classifier.}\label{tab:dropout}
\vspace{-1.5em}
\end{center}
\end{table}

\begin{table}[h]
\small
\begin{center}
\begin{tabular}{|p{25mm}|p{13.5mm}|p{9mm}|p{13.5mm}|} 
\hline
 \rowcolor{gray!50}
 \textbf{Description} & \textbf{Precision} & \textbf{Recall} & \textbf{Accuracy} \\ [0.5ex] 
 \hline
 \textit{pos\_weight} = 2 & 0.7 & 0.71 & 0.72 \\ 
 \hline
 \textit{pos\_weight} = 5 & 0.79 & 0.57 & 0.61 \\ 
 \hline
 \textit{pos\_weight} = 10 & 0.82 & 0.57 & 0.6 \\ 
 \hline
\end{tabular}
\caption{Comparision of performance using different \textit{pos\_weight} values.}\label{tab:tension_pos_weight}
\vspace{-1.5em}
\end{center}
\end{table}

\paragraph{Weight of positive examples} Comparison of using different \textit{pos\_weight} parameter can be found in \autoref{tab:tension_pos_weight}. All experiments consist of 3 linear blocks with dropout equal to 0.4, weight decay equal to 0.001, learning rate equal to 0.0005, and the base model was RoBERTa large. 

\paragraph{Number of linear blocks} The number of linear blocks directly influenced the number of trainable parameters utilised by the model. The base model used was RoBERTa large. After exploring various configurations, the 3 linear blocks perform best as shown in \autoref{tab:layers}.

\begin{table}[h]
\small
\begin{center}
\begin{tabular}{|p{22mm}|p{13.5mm}|p{10mm}|p{13.5mm}|} 
\hline
 \rowcolor{gray!50}
 \textbf{Description} & \textbf{Precision} & \textbf{Recall} & \textbf{Accuracy} \\ [0.5ex] 
 \hline
 N = 3, Dropout = 0.6 & 0.82  &    0.53   &   0.55  \\ 
 \hline
 N = 2, Dropout = 0.6  & 0.80   &   0.61  &    0.64 \\ 
 \hline
 N = 1, Dropout = 0.6 & 0.79  &    0.65 &     0.68 \\ 
 \hline
 N = 3, Dropout = 0.4 & 0.82   &    0.58  &    0.61  \\ 
 \hline
 N = 2, Dropout = 0.4  & 0.80   &   0.61  &    0.64 \\ 
 \hline
 N = 1, Dropout = 0.4 & 0.79  &    0.65 &     0.68 \\ 
 \hline
\end{tabular}
\caption{Comparision of performance for different numbers of layers (N) in the tension classifier.}\label{tab:layers}
\vspace{-1.5em}
\end{center}
\end{table}

\paragraph{Comparing RoBERTa base and RoBERTa large} 

Initially, we assumed that a larger model would be a better choice, as tension is a complex concept. As the last set of experiments, summarised in \autoref{tab:base-large}, we wanted to investigate how many parameters we need to catch the complexity of tension. To do so, we compared the results between RoBERTa base and RoBERTa large as a base model. We found that using the larger model with correct hyperparameters is no better than using the smaller one. 

\begin{table}[h]
\small
\begin{center}
\begin{tabular}{|p{25mm}|p{15mm}|p{10mm}|p{13.5mm}|} 
\hline
 \rowcolor{gray!50}
 \textbf{Description} & \textbf{Precision} & \textbf{Recall} & \textbf{Accuracy} \\ [0.5ex] 
 \hline
 \textit{RoBERTa base}, \textit{pos\_weight} = 5, \textit{do} = 0.4 with pre-fine-tuning & 0.74 & 0.70 & 0.72 \\ 
 \hline
  \textit{RoBERTa base}, \textit{pos\_weight} = 10, \textit{do} = 0.4 with pre-fine-tuning & 0.8 & 0.36 & 0.33 \\ 
 \hline
  \textit{RoBERTa base}, \textit{pos\_weight} = 5, \textit{do} = 0.4 without pre-fine-tuning & 0.79     & 0.34  &  0.30 \\ 
 \hline
   \textit{RoBERTa base}, \textit{pos\_weight} = 5, \textit{do} = 0.6 without pre-fine-tuning & 0.70 &  0.76  & 0.71  \\ 
 \hline
 \textit{RoBERTa base}, \textit{pos\_weight} = 2, \textit{do} = 0.4 with pre-fine-tuning  & 0.7 & 0.71 & 0.72 \\ 
  \hline
 \textit{RoBERTa base}, \textit{pos\_weight} = 10, \textit{do} = 0.4 with pre-fine-tuning & 0.79 & 0.72 & 0.72 \\ 
 \hline

\end{tabular}
\caption{Comparision of performance between RoBERTa base and RoBERTa large as a base model. }\label{tab:base-large}
\vspace{-1.5em}
\end{center}
\end{table}

\paragraph{Results}
The experiments conducted demonstrated that pre-fine-tuning had a positive impact on the results, improving them by 5-10\%. While these improvements were relatively small, they underscore the need to investigate further the influence of different language styles on the model's inference performance. \\ 
We analyse the effect of various hyperparameters on the model's performance. Setting the dropout to 0 resulted in the highest accuracy, although it did not necessarily yield the best precision. A dropout value of 0.4 was chosen as a compromise to balance accuracy and precision. Additionally, we found that setting the \textit{pos\_weight} to 2 improved the overall accuracy. Surprisingly, the findings revealed that employing only one linear block achieved the best recall while maintaining comparable precision and overall accuracy. It proves that a simpler model architecture can effectively capture the relevant features and achieve optimal recall.

\section{Application}
\begin{figure}
    \centering 
    \includegraphics[width=\columnwidth]{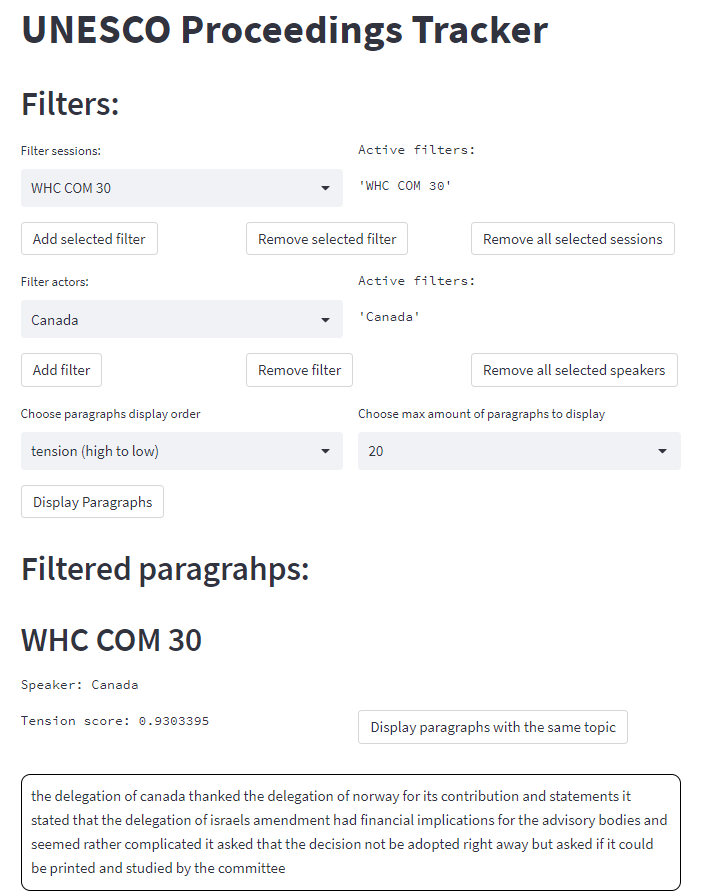}  
    \caption{Sample screen from the application.}
    \label{fig:app_ss}
\end{figure}

We have developed an application specifically designed for researchers of global heritage regimes and UNESCO diplomats to facilitate their search for information within selected speeches. Previously, these individuals had to devote hours to studying extensive summary records to locate relevant fragments for their research or diplomatic practice. However, our application is intended to drastically reduce the time required for this task, enabling users to quickly find the specific statements they need. The application offers the following filtering options for the displayed paragraphs:
\begin{itemize}
    \item \textbf{session}: specifies the sessions from which the paragraphs are displayed.
    \item \textbf{actor}: specifies the speakers of the paragraphs.
\end{itemize}
Furthermore, users can specify the number of paragraphs to be displayed and the preferred order of presentation, either by tension or by date.
In conjunction with each presented paragraph, the application provides additional details, including the speaker's identity, a tension score, and a convenient button that enables users to reveal all paragraphs related to the selected paragraph. All these features are presented in \autoref{fig:app_ss}. 

\section{Related work}\label{sec:related_work}
Recent research on controversy (and, in our case, tension) detection is not broad. We found only one model available publicly, which detects controversy, specifically in the Guardian16 corpus\footnote{\url{https://drive.google.com/file/d/1g6yh77tBgWlgXcKCLULLBVcUt7Xvs1JE/view}} and is described in further detail at~\cite{youngwoo:2019}, where it is additionally stated that a generic document without implicit or explicit topic annotations cannot only rely on inherent topic annotation. The subset of data involving comments has similar issues to other Twitter-based controversy datasets~\cite{twitter_controversy}, as the language used is exceptionally informal and often consists of short sentences. It differs from our dataset, as diplomatic language is significantly more formal. \\
Another popular dataset idea in the controversy classification community contains labelled articles fetched from Wikipedia~\cite{controversy-wikipedia}. The method for building this dataset was first described in~\cite{shiri-dataset} and later expanded upon in a doctoral dissertation~\cite{controversy-phd-dataset}. From that point, researchers build their Wikipedia-derived datasets~\cite{controversy-semantic-approaches} where positive examples were based on Wikipedia's \emph{List of controversial issues}\footnote{\url{https://en.wikipedia.org/wiki/Wikipedia:List\_of\_controversial\_issues}} for their own need, but rarely making the produced textual data public. IBM researchers used original description~\cite{shiri-dataset} to produce a downloadable dataset named \emph{dataset\_ii.csv}\footnote{\url{https://research.ibm.com/haifa/dept/vst/debating\_data.shtml}}.\\
In addition, they extracted 3561 concepts, crowd-annotated later, from Wikipedia pages under edit protection, assuming that many of these would be controversial. This dataset was named \emph{dataset\_iii.csv}\footnote{\url{https://research.ibm.com/haifa/dept/vst/debating\_data.shtml}}. The average pairwise Cohen Kappa agreement on this task was 0.532.
\autoref{tab:other_datasets} illustrates each dataset's negative (0s) and positive samples (1s) and thus shows an imbalance between classes that are needed to address. It's worth noting that this set of textual data was never used together in controversy detection research. 

\begin{table}[h]
\small
\begin{center}
\begin{tabular}{|p{30mm}|c|c|c|} 
\hline
 \rowcolor{gray!50}
 \textbf{Name of the dataset} & \textbf{Non-tension}  & \textbf{Tension} & \textbf{Total} \\ [0.5ex] 
 \hline
 Guardian & 439 & 281 & 720 \\ 
 \hline
 Wiki ii & 608 & 605 & 1213 \\
 \hline
 Wiki iii & 2720 & 841 & 3561 \\
 \hline
 Guardian + Wiki & 3767 & 1727 & 5494 \\
 \hline
 Guardian + Wiki + Comments & 100435 & 172093 & 272528 \\ [1ex] 
 \hline
\end{tabular}
\end{center}
    \caption{Number of positive and negative samples for each dataset.}
    \label{tab:other_datasets}
\end{table}

\section{Conclusions}
During our research, we have successfully developed a pioneering tool for the computational analysis of UNESCO World Heritage Convention (WHC) and Intangible Culture Heritage Convention (ICHC) proceedings. This tool encompasses many features and functionalities, catering to the diverse needs of diplomats and political scientists analysing these important textual resources.\\ 
To achieve the primary goal of detecting tensions within the text, we harnessed the power of pre-trained language models and enhanced them by incorporating additional layers. By doing so, we have successfully created a classifier that operates in the complex and multifaceted domain of political science, specifically within the realm of UNESCO proceedings.\\
The development of our tool marks a significant advancement in the field, providing researchers and practitioners with a robust solution for computational analysis and exploration of tensions within these important discourse contexts.\\
Our findings contribute to understanding tension in a specific domain and provide valuable insights for further research in related areas. 
\section{Limitations and future work}
While the proposed methodology for analysing diplomatic documents presented in this paper offers significant contributions to the field, it is important to acknowledge certain limitations and potential areas for improvement. These limitations include:
\begin{itemize}
    \item \textbf{Scalability}: Annotating controversies is time-consuming and resource-intensive. Creating a large annotated dataset requires significant effort and expertise. As a result, the current dataset size may not be sufficient to capture the full complexity and variability of tensions. Future research should aim to overcome scalability challenges and develop strategies for efficiently creating larger annotated datasets, for example, by adding more active learning loops. 
    \item \textbf{Generalisation to Other Political Organizations}: The proposed methodology's effectiveness in detecting tensions in other political organisations is uncertain. Different political organisations often have distinct ideologies, rhetoric, and controversies that may not align with the training data. The model may not effectively capture tensions' unique characteristics and dynamics in diverse political contexts. Our model is based only on the UNESCO dataset, but we suspect it can represent the language political scientists use well. We plan to create a model fine-tuned on datasets containing diplomats' speeches that can be used in the diplomatic language in NLP tasks.  
    \item \textbf{Variability in Speaker References}: Identifying speakers solely through regular expressions may be challenging when multiple ways of referring individuals or groups exist. Speakers can be referred to using various forms, such as names, pronouns, titles, or descriptions. Regular expressions alone may not capture all possible variations and may lead to inaccurate or incomplete speaker detection. Developing a robust tool for detecting speaking actors in any reported speech data would enhance detection and facilitate generalisation to other problems.
    \item \textbf{Extending range of tension:} In this work, we've focused only on binary classification of tension. However, in real-world scenarios, tension is often a nuanced and multi-dimensional concept that cannot be adequately captured by a simple binary classification. Future work could explore the possibility of extending the range of tension by considering a more fine-grained approach. 
\end{itemize}

\section*{Acknowledgements}
All experiments were performed using the Entropy cluster funded by NVIDIA, Intel, the Polish National Science Center grant UMO-2017/26/E/ST6/00622 and ERC Starting Grant TOTAL. This article describes a Team Programming Project completed at the University of Warsaw in the academic year 22/23. Hanna Schreiber and Bartosz Pieliński wish to acknowledge that their contribution to this paper was carried out within the framework of the research grant Sonata 15, "Between the heritage of the world and the heritage of humanity: researching international heritage regimes through the prism of Elinor Ostrom's IAD framework," 2019/35/D/HS5/04247 financed by the National Science Centre (Poland).

\bibliography{custom}
\bibliographystyle{acl_natbib}

\appendix

\section{Details about splitting text into paragraphs}\label{appendix:splitting_text}
The primary regular expression was designed to identify sentence beginnings following vertical breaks. By carefully considering exceptions, such as page endings, we achieved a high success rate in locating the majority of paragraphs. Additionally, we dedicated efforts to creating multiple specialised regular expressions capable of detecting unique patterns observed in specific summary records, such as bullet points or slides.

\section{Details about expert-labelling}\label{appendix:whc_ich_labelled}
 Our experts labelled two sessions: 35 WHC ordinary session\footnote{\url{https://whc.unesco.org/en/sessions/35COM}}, which encompassed 654 paragraphs, and 12 ICHC ordinary session\footnote{\url{https://ich.unesco.org/en/12com}}, which consisted of 616 paragraphs. Initially, there was a notable discrepancy in their annotations, primarily due to the lack of strict guidelines for labelling the positive class. However, once stricter guidelines were established, the distribution between paragraphs containing tension and those that did not change, as indicated in \autoref{table:whc-ich-imbalance}.\\
The row \emph{Consistent annotation from beginning} statistic provides valuable insights into the level of agreement between our expert annotators regarding the presence or absence of tension in the annotated paragraphs. \\
Our analysis revealed that, in the ICHC dataset, 39 paragraphs received a unanimous label indicating the presence of tension. Similarly, in the WHC dataset, 17 paragraphs were consistently identified by both annotators as containing tension. \\
After the data was labelled and before the conflicts were resolved, we computed the score of their annotations, called the Cohen kappa score~\cite{kappa}, which measures the compatibility of two annotators in categorical classification. The Cohen kappa score for paragraphs from 12 ICHC COM was equal to: 0.2205, and for paragraphs from 35, WHC COM: 0.1148. The low score was the effect of an insufficient description of tension in annotation guidelines. Together with domain experts we've fixed the guidelines. Its final version is available in \autoref{sec:annotation_guidelines}.\\
Initially, 404 paragraphs from WHC and ICHC datasets exhibited complexities in achieving unanimous annotation agreement. However, through rigorous examination and expert discourse, a consensus was reached, and additional 166 paragraphs from the 35 WHC COM dataset and 99 paragraphs from the 12 ICHC COM dataset were labelled as positive.\\ 
These findings highlight the inherent challenges associated with the annotation process and underscore the significance of expert discussions and consensus-building to ensure the accurate classification of tension within the analysed paragraphs.
\begin{table}[H]
\small
\begin{center}
\begin{tabular}{|p{40mm}|c|c|c|}
 \hline
 \rowcolor{gray!50}
 Description of a subset & 0s & 1s & Total \\ [0.5ex] 
 \hline
 Full 35 WHC & 471 & 183 & 654 \\
 \hline
 Full 12 ICHC & 478 & 138  & 616 \\ 
 \hline
 Full 35 WHC and 12 ICHC & 949 & 321 & 1270 \\ 
 \hline
 Consistent annotation of full 35 WHC and 12 ICHC after the first annotation stage & 810 & 56 & 866 \\
 \hline
 35 WHC without introduction & 451  & 183 & 634 \\
 \hline
 12 ICHC without introduction & 458  & 138 & 596  \\
 \hline
 35 WHC and 12 ICHC without introduction & 909  & 321 & 1230  \\
 \hline
\end{tabular}
\end{center}
\caption{Details about the annotation of datasets.}
\label{table:whc-ich-imbalance}
\end{table} 

\subsection{Annotation guidelines}\label{sec:annotation_guidelines}

Annotation was done by two researchers and co-authors of this paper. The first annotator was a political scientist with extensive expertise in text analysis (Bartosz Pieliński). The second one was international affairs researcher and long-time UNESCO cultural heritage expert (Hanna Schreiber). Each annotator was presented with the annotation guidelines as stated below.

    \paragraph{Introduction} In this task, you aim to detect tensions in UNESCO Summary Records, transcriptions from UNESCO sessions. Tensions refer to controversial issues rooted in disagreements related to states' interests and values. The annotation task involves classifying paragraphs as either indicating tension (1) or not indicating it (0). You should follow the guidelines below to ensure consistency and accuracy in annotation process.

    \paragraph{Annotation schema}
    \begin{itemize}
        \item \textbf{Tension} Mark a paragraph as indicating tension if (1) there is a controversy between participants of a discussion, and (2) the controversy relates to the interests or values of at least one of the actors taking part in the discussion.
        \item \textbf{No Tension} Mark a paragraph as not indicating tension if (1) there is no controversy between participants of a discussion or if (2)  there is a controversy, but it is not related to the interests or values of at least one of the actors taking part in the discussion.
    \end{itemize}

    \paragraph{Document Segments} Each document is splitted into paragraphs. They may vary in length, ranging from a single word to several sentences. You should read and analyse each segment to determine its classification based on the provided annotation schema.

    \paragraph{Annotator Instructions}
    \begin{itemize}
        \item Familiarise yourself with the topic of the research and the context of diplomacy documents.
        \item Focus on identifying any indications of tension, disagreement, or conflicting positions within the segment.
        \item Make the annotation judgment based solely on the content of the segment itself; do not consider information from other parts of the document or external sources.
        \item Use your best judgment and avoid making assumptions or inferences beyond what is explicitly stated in the text.
        \item If you encounter ambiguous segments or are uncertain about the classification, mark them for review, and consult with the research team.
    \end{itemize}

    \paragraph{Annotation Process}
    \begin{itemize}
        \item Use the annotation tool provided by the research team to mark each segment as tense or non-tense.
        \item Pay close attention to sentence boundaries and ensure the annotation accurately represents the segment's overall meaning.
        \item If a document segment contains a mix of tense and non-tense elements, consider the dominant tone and classify it accordingly.
        \item Do not modify the original document or alter the text in any way during the annotation process.
    \end{itemize}

    \paragraph{Inter-Annotator Agreement (IAA)} To ensure the reliability of the annotations, at least two annotators will independently review each document segment. The research team will provide a guideline for handling cases of ambiguous or challenging segments to promote consistent annotations.

    \paragraph{Confidentiality and Data Handling} Treat all research documents and data as confidential and only use them for the purpose of this research project. Do not share or discuss any document content or results with unauthorised individuals or outside the research team.

    \paragraph{Annotation Completion and Review} Inform the research team once you have completed the annotation task. Participate in review meetings with the research team to address any questions, concerns, or discrepancies in the annotations.

    By following these guidelines, you can contribute to the creation of a reliable dataset for detecting tensions in diplomacy documents, facilitating the research's success and impact.

\section{Removed phrases}\label{appendix:removed_phrases}
The full list of hand-picked phrases we removed from consideration during topic modelling is provided below. Moreover, we omitted all descriptions of nationalities and country names. \\

\begin{minipage}{0.45\columnwidth}
\begin{itemize}
    \item chairperson
    \item committee
    \item cultural
    \item delegate
    \item delegation 
    \item heritage
    \item iccrom 
    \item icomos
    \item iucn
    \item lesion
\end{itemize}
\end{minipage}
\hfill
\begin{minipage}{0.45\columnwidth}
\begin{itemize}
    \item outstanding
    \item party
    \item property
    \item rapporteur
    \item representation
    \item session
    \item representative 
    \item site 
    \item state
    \item world
\end{itemize}
\end{minipage}

\newpage
\section{ICH speaker extraction}\label{app:ich}

\begin{figure}[ht]
    \centering
    \includegraphics[width=\columnwidth]{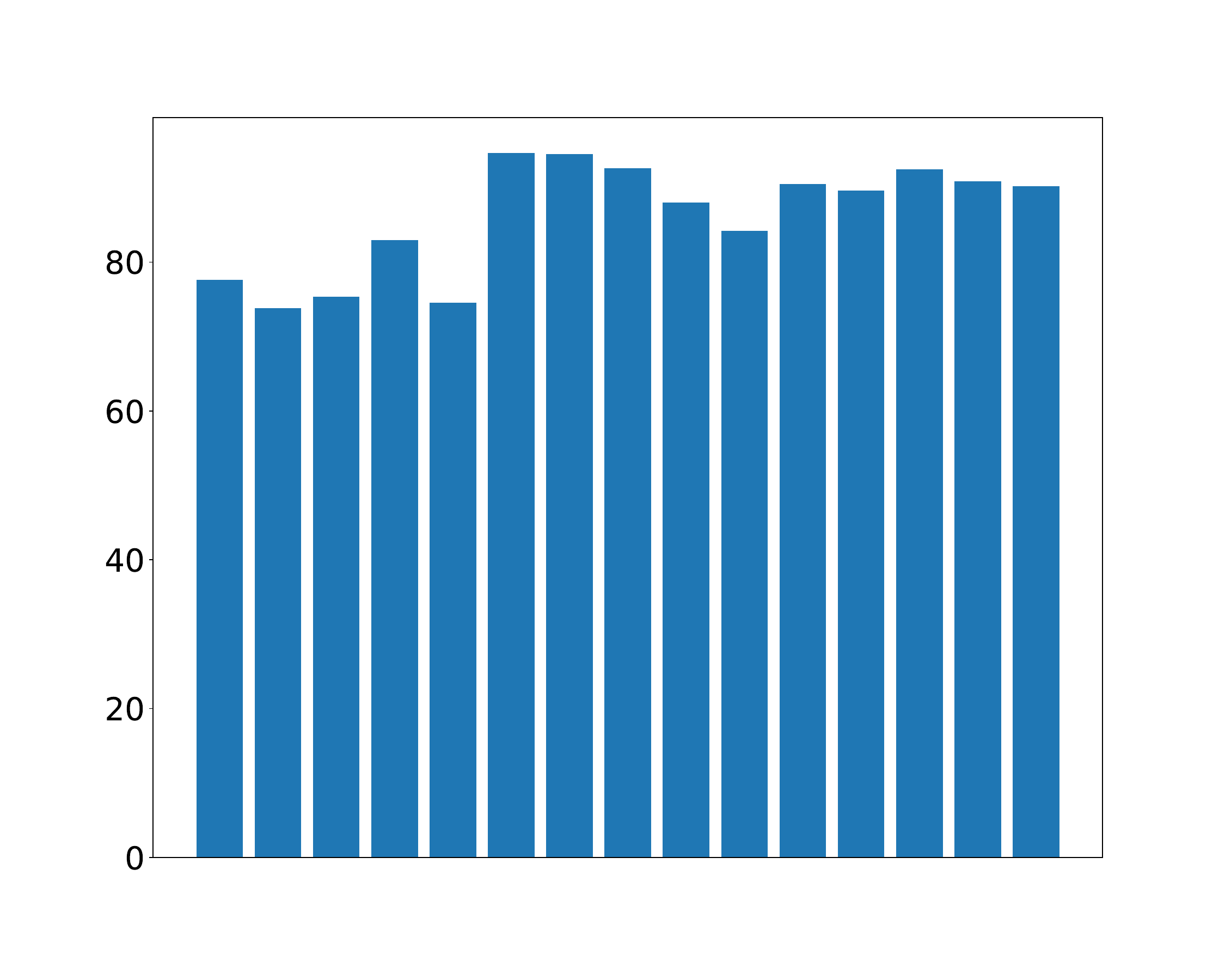}

    \caption{Percentage of paragraphs with identified speakers in each ordinary ICH session.}
    \label{fig:ich_speakers}
\end{figure}

\end{document}